# Texygen: A Benchmarking Platform for Text Generation Models

Yaoming Zhu[†], Sidi Lu[†], Lei Zheng[†], Jiaxian Guo[†], Weinan Zhang[†], Jun Wang[‡], Yong Yu[†*]

[†]Shanghai Jiao Tong University, [‡]University College London
{ymzhu,steve_lu,wnzhang}@apex.sjtu.edu.cn

## ABSTRACT

We introduce *Texygen*, a benchmarking platform to support research on open-domain text generation models. Texygen has not only implemented a majority of text generation models, but also covered a set of metrics that evaluate the diversity, the quality and the consistency of the generated texts. The Texygen platform could help standardize the research on text generation and facilitate the sharing of fine-tuned open-source implementations among researchers for their work. As a consequence, this would help in improving the reproductivity and reliability of future research work in text generation.

## 1 INTRODUCTION

The open-domain text generation problem aims at modeling the sequential generation of discrete tokens. It has rich real-world applications, including, but not limited to, machine translation [2], AI chat bots [9], image captioning [15], question answering and information retrieval [13]. While we have witnessed various implementations of its practical use, the fundamental research on text generative models has also made significant progress. Notably, since maximum likelihood estimation (MLE) [11] is not a perfect objective function for this sequential generation problem (due to "exposure bias" explained in [6]), researchers have been looking for alternative optimization methods and objective functions.

The success of Generative Adversarial Network (GAN) [4] has inspired people to investigate adversarial training over textual, discrete data. Sequence Generative Adversarial Network, a.k.a. SeqGAN, for example, is one of the very early attempts at applying REINFORCE algorithm [14] to solving the discrete optimization of GAN objective. Since then, many methods about improving SeqGAN are proposed to further develop SeqGAN in many aspects; examples include gradient vanishing (MaliGAN [3], RankGAN [10], BRA in LeakGAN [5]), long-term robustness (LeakGAN).

There are three main challenges on evaluating sequential generation of textual data. First, the criterion of what is a good text generation model is still unclear. Although researchers have developed some metrics such as perplexity [7], oracle generated log-likelihood [16], Turing-test-based human scores and the BLEU metric [12], there is no single metric that is comprehensive enough for measuring the performance of a text generation model. Thus, evaluation over multiple metrics is required to draw definitive answers. Second, there is no obligation for researchers to make their source code publicly available, thus making reproducing the reported experimental results difficult. Third, text generation suffers from a so-called quality-diversity tradeoff problem, i.e., to-some-extent shrink to a restricted output pattern with mode collapse, which, on the other hand, causes researchers to release those models which only actually adjust the tradeoff balance. To our knowledge, there is not a good metric focused on the diversity of the text generation. Thus, there is a significant need for a reliable platform that provides a thorough evaluation of the existing text generation models and facilitate the development of new ones in a common framework.

In this paper, we release Texygen[1], a fully open-sourced benchmarking platform for text generation models. Texygen not only includes a majority of the baseline models, but also maintains a variety of metrics that evaluates the diversity, quality and the consistency of the generated texts. With these metrics, we can have a much more comprehensive study of different text generation models. We hope this platform could help the progress of standardizing the research on text generation, increase the reproducibility of research work in this field, and encourage higher-level applications.

## 2 THE TEXYGEN PLATFORM

Texygen provides a standard top-to-down multi-dimensional evaluation system for text generation models. Currently, Texygen consists of two elements: well-trained baseline models and automatically computable evaluation metrics. Texygen also provides the open source repository of the platform, in which researchers can find the specification and manual of APIs for implementing their models for Texygen to evaluate.

### 2.1 Baseline Models

In the current version of Texygen, we implement various likelihood-based models such as vanilla MLE language model, SeqGAN [16], MaliGAN [3], RankGAN [10], TextGAN (feature matching) [17], GSGAN (GAN with Gumbel Softmax trick) [8] and LeakGAN [5]. These baseline models contain supervised likelihood-based methods with and without tricks, adversarial methods and hierarchical methods. Although more models will be added, we believe the current coverage in the collection is sufficient for good comparison of any arbitrary new model. Here we briefly introduce these models.

**Vanilla MLE.** Given a sequence piece $s_t = [x_0, x_1, ..., x_{t-1}]$ and the next token to be sampled from the model $x_t \sim \pi_\theta(x|s_t)$, a vanilla MLE language model [11] adopts an explicit likelihood-based modeling of language, with the form:

$$\max_\theta \sum_{\mathbf{x}} \sum_t \log \pi_\theta(x_t|s_t)$$

where **x** iterates over training data sentences and $t$ iterates over the token sequence of each sentence. By maximizing the likelihood estimation, MLE manages to have an estimation of the generation procedure.

**SeqGAN.** SeqGAN [16] adopts a discriminative model that is trained to minimize the binary classification loss between real texts and generated texts. Meanwhile, besides the pretraining procedure that follows MLE metric, the generator uses the REINFORCE algorithm to optimize the GAN objective

$$\min_{\phi} - \mathbb{E}_{Y \sim p_{\text{data}}} \left[ \log D_{\phi}(Y) \right] - \mathbb{E}_{Y \sim G_{\theta}} \left[ \log(1 - D_{\phi}(Y)) \right].$$

---

[*]W. Zhang is the corresponding author.

[1]https://github.com/geek-ai/Texygen

For the sake of variance reduction, SeqGAN uses Monte Carlo search to compute the Q-value for generating each token.

**MaliGAN.** The basic structure of MaliGAN [3] follows that of the SeqGAN. To stablize the training and alleviate the gradient saturating problem, MaliGAN rescales the reward in a batch with size $m$ by

$$r_D(x_i)' = \frac{r_D(x_i)}{\sum_{f=1}^{m} r_D(x_f)} - b,$$

where $r_D(\cdot)$ is the reward function from the discriminator, $b$ is the moving average of $r_D(\cdot)$ as the baseline.

**RankGAN.** RankGAN [10] replaces the discriminator of SeqGAN with a ranker $R_{\phi}$, which optimizes the ranking loss

$$L_{\phi} = \mathbb{E}_{s \sim p_{\text{data}}} \left[ \log R_{\phi}(s|U, C^-) \right] - \mathbb{E}_{s \sim G_{\theta}} \left[ \log R_{\phi}(s|U, C^+) \right],$$

where

$$R_{\phi}(s|U, C) = \log \left( \frac{\exp(\gamma \alpha(s|u))}{\sum_{s' \in C} \exp(\gamma \alpha(s'|u))} \right)$$

$$\alpha(s|u) = \cos(y_s, y_u)$$

**GSGAN.** Gumbel Softmax trick is a reparametrization trick used to replace the multinomial stochastic sampling in text generation [8]. It claims that

$$\arg\max \left[ \text{softmax}(h + g) \right] \sim \text{softmax}(h),$$

where $g$ is a Gumbel distribution with zero mean and unit variance. Note that since this process is differentiable, thus backpropagation can be directly applied to optimize the GAN objective.

**TextGAN.** Adversarial feature matching for text generation [17] proposes a method that optimizes the MMD loss, which is the reconstructed feature distance, by adding a reconstruction term in the objective, i.e.,

$$L_{\text{recon}} = \|z - \hat{z}\|.$$

**LeakGAN.** LeakGAN [5] is a hierarchical reinforcement learning framework with two modules called Manager and Worker respectively. In general, the Manager learns to generate a sequence of subgoals for the sequence and the Worker learns to fulfill it.

## 2.2 Metrics

Texygen implements five text generation metrics so far, covering various aspects as categorized below. It also provides user-friendly APIs to retrieve results of their own models and generated text.

### 2.2.1 Document Similarity based Metrics.
The most intuitive measurement of generated documents quality is how the documents resemble the natural language, or, the training dataset.

**BLEU.** BLEU [12] is a widely used metric evaluating the word similarity between sentences or documents.

**EmbSim.** Inspired by BLEU, we propose EmbSim to evaluating the similarity between two documents, whose name stands for "embedding similarity". Instead of comparing sentences words by words, we compare the word embeddings.

First, word embedding is evaluated on real data using a skip-gram model. For each word embedding, we compute its cosine distance with the other words, and then formulate it as a matrix $W$, where $W_{i,j} = \cos(e_i, e_j)$ with $e_i$, $e_j$ being the word embeddings of the word $i$ and $j$ from real data. We call $W$ the similarity matrix of real data.

Similarly, we get the similarity matrix $W'$ of generated data, where $W'_{i,j} = \cos(e'_i, e'_j)$ with $e'_i$ and $e'_j$ being the word embedding of the word $i$ and $j$ from generated data using the same skip-gram model.

The EmbSim is defined as

$$\text{EmbSim} = \log \left( \sum_{i=1}^{N} \cos(W'_i, W_i)/N \right)$$

where $N$ is the total number of words, and $W_i$ and $W'_i$ denote the $i$-th column of $W$ and $W'$ respectively.

### 2.2.2 Likelihood-based Metrics.
Based on MLE, which aims at minimizing the cross-entropy between the true data distribution $p$ and the generated data distribution $q$ from the model, we can design metrics to evaluate how good data and model is fitted by measuring the likelihood. These models require details about not only data but the model as well.

**NLL-oracle.** Negative log-likelihood (NLL) is originally introduced in SeqGAN [16], which is specifically applied on synthetic data experiment and tells how good the generated data is fitted by the oracle language model. In NLL$_{\text{oracle}}$, a randomly initialized LSTM is regarded as a true model, i.e., the oracle. Text generation models need to minimize average negative log-likelihood of generate data on oracle LSTM, i.e. $\mathbb{E}_{x \sim q} \log p(x)$, where $x$ denotes the generated data.

Since an LSTM is regarded as a true model, the metric can calculate the average loss on every sentence, word by word

$$\text{NLL}_{\text{oracle}} = -\mathbb{E}_{Y_{1:T} \sim G_{\theta}} [\sum_{t=1}^{T} \log(G_{\text{oracle}}(y_t|Y_{1:t-1}))],$$

where $G_{\text{oracle}}$ denotes the oracle LSTM, and $G_{\theta}$ denotes the generative model.

**NLL-test.** We propose NLL$_{\text{test}}$, a simple metric evaluating the model's capacity to fit real test data, which is dual to NLL$_{\text{oracle}}$.

$$\text{NLL}_{\text{test}} = -\mathbb{E}_{Y_{1:T} \sim G_{\text{real}}} [\sum_{t=1}^{T} \log(G_{\theta}(y_t|Y_{1:t-1}))],$$

where $G_{\text{real}}$ denotes the distribution of real data.

NLL$_{\text{test}}$ can only be applied to autoregressive generator like RNN since $G_{\theta}(y_t|Y_{1:t-1})$ is involved to calculate the likelihood of certain word based on previous ones given a generator.

### 2.2.3 Divergence based Metrics.
GAN models often suffer from mode collapse problems, which lead to generator collapsing to produce only a single sample or a small family of very similar samples. Thus, in open-domain text generation tasks, we include metrics that encourage to generate more diverse patterns.

**Self-BLEU.** We propose Self-BLEU, a metric to evaluate the diversity of the generated data. Since BLEU aims to assess how similar two sentences are, it can also be used to evaluate how one sentence resembles the rest in a generated collection. Regarding one sentence as hypothesis and the others as reference, we can calculate BLEU score for every generated sentence, and define the average BLEU score to be the Self-BLEU of the document.

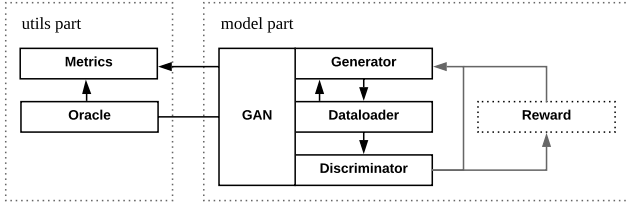

Figure 1: Texygen architecture.

A higher Self-BLEU score implies less diversity of the document, and more serious mode collapse of the GAN model.

## 3 PLATFORM ARCHITECTURE

Texygen is implemented over TensorFlow [1]. As shown in Fig. 1, the system consists of two parts with three major classes, highly decoupled with each other, and easy for customization.

In the *utils* part, we provide user Metrics class and Oracle class. The former has three subclasses designed for calculating BLEU score, NLL loss and EmbSim, while the latter one enables user to initialize three different types of Oracle: LSTM-based, GRU-based and SRU-based. The default oracle is LSTM.

In the model part, we enable users to begin the training process by only interacting with the GAN class (as a major class) without concerning about the classes for the generator, the discriminator and the reward (for RL-based GANs). Texygen also provides two different types of training processes in the GAN class: synthetic data training and real data training. The former one uses the oracle LSTM to generate data, while the latter one uses real-world datasets (e.g., COCO image caption[2]).

## 4 EXPERIMENT

### 4.1 Training Setting

**Data.** In our synthetic data training, the total number of words is set to be 5,000 and the sentence length is set to be 20, and the oracle will generate 10,000 sentences. In the real data training, we select 20,000 sentence from the image COCO captions, with half of them as the training set, the rest as the test set.

**GAN Setting.** The default initial parameter of all generator follows a Gaussian distribution $\mathcal{N}(0, 1)$. We use MLE training as the pretraining process for all baseline models except GSGAN, which requires no pretraining. In pretraining, we first train 80 epochs for a generator, and then 80 epochs for a discriminator. The adversarial training comes next. In each adversarial epoch, we update the generator once and then update the discriminator for 15 mini-batch gradients. In LeakGAN, after every 10 adversarial epochs, there are 5 MLE training epochs for both the generator and the discriminator. The total number of adversarial training epochs is 100.

**Evaluation Metrics.** $\text{NLL}_{\text{oracle}}$ and $\text{NLL}_{\text{test}}$ are applied to synthetic data training. Since the oracle LSTM cannot generate words with semantic meaning, we do not calculate the BLEU score or EmbSim on synthetic data. On the other hand, the BLEU score, the Self-BLEU score and EmbSim are applied to real data training.

[2]http://cocodataset.org/

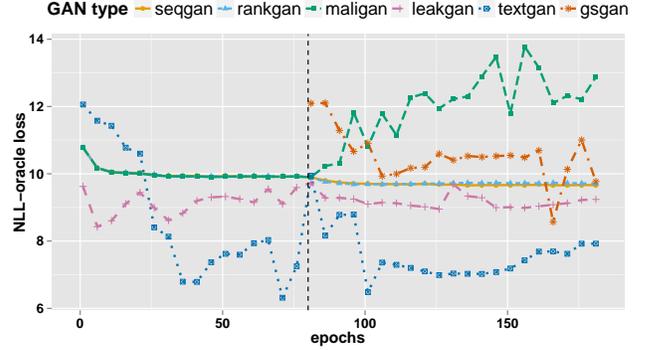

Figure 2: NLL-oracle loss comparison throughout training.

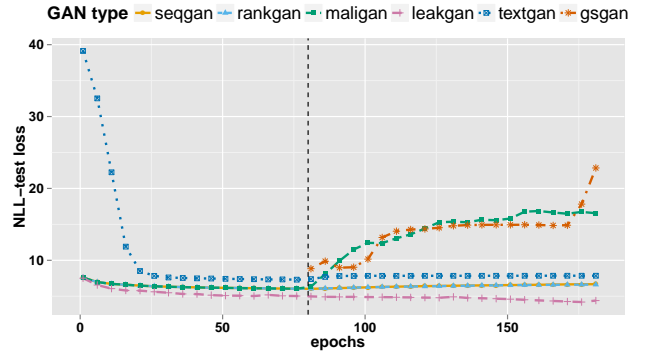

Figure 3: NLL-test loss comparison throughout training.

### 4.2 Synthetic Data Experiment

The training curves of $\text{NLL}_{\text{oracle}}$ and $\text{NLL}_{\text{test}}$ are depicted in Figs. 2 and 3 respectively. LeakGAN converges more quickly and achieves good performance on these two metrics. TextGAN gets best $\text{NLL}_{\text{oracle}}$ results, while gets worst $\text{NLL}_{\text{test}}$ performance on pretraining epochs. Due to model similarity, SeqGAN, MaliGAN and RankGAN have almost identical curves until adversarial epochs, after which MaliGAN becomes less competitive compared to the other two.

### 4.3 Real Data Experiment

The training curves of EmbSim is depicted in Fig. 4. LeakGAN achieves very high similarity at the beginning epochs while TextGAN has rather slow improvement. All GAN models achieve the best results on pretraining steps. Once the adversarial training starts, only LeakGAN still maintains its EmbSim score, while other baseline models' EmbSim scores decrease compared to the pretraining epochs.

In this part, GSGAN is excluded, since it fails to generate any sentences with semantics in our experiment. The generated instances can be accessed from Texygen webpage.

The BLEU score on training data, test data is shown in Tables 1 and 2 respectively. LeakGAN outperforms other baseline models on the metric, and its performance on test dataset shows it has rather

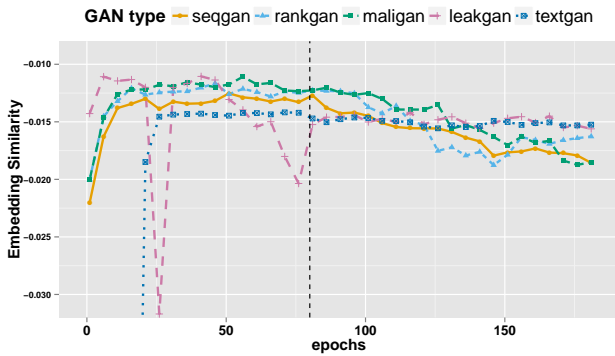

Figure 4: EmbSim comparison throughout training.

Table 1: BLEU score on training data

|        | SeqGAN | MaliGAN | RankGAN |
|--------|--------|---------|---------|
| BLEU-2 | 0.917  | 0.887   | **0.937** |
| BLEU-3 | 0.747  | 0.697   | 0.799   |
| BLEU-4 | 0.530  | 0.482   | 0.601   |
| BLEU-5 | 0.348  | 0.312   | 0.414   |
|        | LeakGAN | TextGAN | MLE    |
| BLEU-2 | 0.926  | 0.650   | 0.921  |
| BLEU-3 | **0.816** | 0.645 | 0.768  |
| BLEU-4 | **0.660** | 0.596 | 0.570  |
| BLEU-5 | 0.470  | **0.523** | 0.392 |

Table 2: BLEU score on test data

|        | SeqGAN | MaliGAN | RankGAN |
|--------|--------|---------|---------|
| BLEU-2 | 0.745  | 0.673   | 0.743   |
| BLEU-3 | 0.498  | 0.432   | 0.467   |
| BLEU-4 | 0.294  | 0.257   | 0.264   |
| BLEU-5 | 0.180  | 0.159   | 0.156   |
|        | LeakGAN | TextGAN | MLE    |
| BLEU-2 | **0.746** | 0.593 | 0.731  |
| BLEU-3 | **0.528** | 0.463 | 0.497  |
| BLEU-4 | **0.355** | 0.277 | 0.305  |
| BLEU-5 | **0.230** | 0.207 | 0.189  |

Table 3: Self-BLEU score

|        | SeqGAN | MaliGAN | RankGAN |
|--------|--------|---------|---------|
| BLEU-2 | 0.950  | 0.918   | 0.959   |
| BLEU-3 | 0.840  | 0.781   | 0.882   |
| BLEU-4 | 0.670  | 0.606   | 0.762   |
| BLEU-5 | 0.489  | 0.437   | 0.618   |
|        | LeakGAN | TextGAN | MLE    |
| BLEU-2 | 0.966  | 0.942   | **0.916** |
| BLEU-3 | 0.913  | 0.931   | **0.769** |
| BLEU-4 | 0.848  | 0.804   | **0.583** |
| BLEU-5 | 0.780  | 0.746   | **0.408** |

good generalization capacity. MaliGAN has the lowest BLEU score among all models.

The Self-BLEU score is shown in Table 3. It is clear that all models generate less diverse documents compared to original training data. LeakGAN and TextGAN suffer more serious mode collapse problem compared to the other models, while MLE and MaliGAN can generate documents with the highest diversity.

More detailed empirical study will be made available on Texygen project webpage. For instance, LeakGAN tends to generate longer sentences, while TextGAN is prone to generating short sentences.

## 5 CONCLUSION AND FUTURE WORK

Texygen is a text generation benchmarking platform enabling researchers to evaluate their own models and compare them with existing baseline models fairly and conveniently from different perspectives. Texygen has already designed and implemented various evaluation metrics in order to provide a comprehensive benchmark.

We also discovered that not all metrics in NLP are suitable for text generation. For instances, context free grammar (CFG) is a widely used metric on text grammar analysis, and has been used as a metric in some related work [8]. However, in practice, we found that it cannot distinguish different models and is even prone to favoring ones with more severe mode collapse, as these models may only learn a few grammars. For the future work of this project, we will keep updating new models and designing novel metrics for better benchmarking the text generation tasks.